\documentclass[runningheads]{llncs}
\usepackage[T1]{fontenc}
\usepackage{graphicx,verbatim}

\usepackage{amsmath, amssymb}
\usepackage{mathrsfs}
\usepackage{bm}
\usepackage[capitalize]{cleveref}
\usepackage{kotex}
\usepackage{lscape}
\usepackage{xcolor}
\usepackage{lipsum}
\usepackage{subcaption}
\usepackage{multirow}
\usepackage{float} 
\usepackage{graphicx}
\newcommand{\hide}[1]{}
\usepackage[normalem]{ulem}
\useunder{\uline}{\ul}{}
\usepackage{adjustbox}

\begin{document}

\title{CLIP-KOA: Enhancing Knee Osteoarthritis Diagnosis with Multi-Modal Learning and Symmetry-Aware Loss Functions}

\author{Yejin Jeong, Donghun Lee\thanks{Corresponding author}}
\authorrunning{Jeong and Lee}
\titlerunning{CLIP-KOA}
\institute{
Department of Mathematics, Korea University, Seoul, 02841, Republic of Korea\\
\email{\{yejin\_mds, holy\}@korea.ac.kr}
}
    
\maketitle

\begin{abstract}
Knee osteoarthritis (KOA) is a universal chronic musculoskeletal disorders worldwide, making early diagnosis crucial.
Currently, the Kellgren and Lawrence (KL) grading system is widely used to assess KOA severity. However, its high inter-observer variability and subjectivity hinder diagnostic consistency. To address these limitations, automated diagnostic techniques using deep learning have been actively explored in recent years.
In this study, we propose a CLIP-based framework (CLIP-KOA) to enhance the consistency and reliability of KOA grade prediction. To achieve this, we introduce a learning approach that integrates image and text information and incorporate Symmetry Loss and Consistency Loss to ensure prediction consistency between the original and flipped images.
CLIP-KOA achieves state-of-the-art accuracy of 71.86\% on KOA severity prediction task, and ablation studies show that CLIP-KOA has 2.36\% improvement in accuracy over the standard CLIP model due to our contribution.
This study shows a novel direction for data-driven medical prediction not only to improve reliability of fine-grained diagnosis and but also to explore multimodal methods for medical image analysis.
Our code is available at https://github.com/anonymized-link.


\end{abstract}

\section{Introduction}
Osteoarthritis (OA) is the leading chronic musculoskeletal disease worldwide, affecting approximately 15\% of the population, including around 43 million people in the United States \cite{No_9,No_27}. Knee OA, in particular, is a leading cause of disability and is characterized by a progression from intermittent pain during weight-bearing activities to persistent, chronic pain \cite{No_10}. OA is not merely a degenerative change in articular cartilage but a complex condition that affects the entire joint. Therefore, early diagnosis and tailored treatment strategies are essential for its effective prevention and management.

One of the most widely used methods for diagnosing OA is the Kellgren \& Lawrence (KL) grading system, a radiological osteoarthritis assessment framework proposed in 1957 \cite{No_12}. It evaluates OA severity on a scale from 0 to 4 based on radiological (X-ray) images. However, the KL grading system relies heavily on the physician’s subjective judgment, leading to limitations that can result in variability depending on the physician's experience \cite{No_13}.
In fact, the inter-rater reliability of the KL system has been reported to range from 0.54 to 0.79, with adjacent grades, such as KL 1 and KL 2, being particularly prone to misclassification due to their subtle differences \cite{No_14,No_16}. To address these limitations, automated OA diagnosis using deep learning has been actively explored, highlighting the growing need for more objective and quantitative evaluation methods.

Existing studies have employed various techniques for osteoarthritis(OA) diagnosis, with a particular focus on the KL grading system. However, this system has significant limitations due to high inter-observer variability and reliance on subjective assessment. To address these issues, automated KOA diagnosis systems based on deep learning and machine learning have been actively explored \cite{No_15}.
Some studies have utilized multi-class classification approaches to predict KL grades \cite{No_1,No_4,No_5}, while others have applied ordinal regression techniques to account for the ordinal nature of KL grading \cite{No_2}.

Additionally, Medical Vision-Language Pretraining (VLP) has emerged as a key approach to mitigating medical data sparsity. It has been actively explored to enhance performance in a range of medical image analysis tasks \cite{No_18}. Contrastive Language-Image Pretraining (CLIP) models \cite{No_8} have demonstrated strong generalization capabilities across various downstream vision tasks. They have gained significant attention for their high classification accuracy, particularly in imbalanced or stylistically diverse datasets.
Recent studies have explored the application of CLIP's pretraining techniques to the medical imaging domain, leading to various reported use cases \cite{No_19}. Notable examples include Diabetic Retinopathy analysis \cite{No_7}, Histopathology image interpretation \cite{No_20}, and PubMedCLIP, which incorporates Pathology, Blood, and Breast imaging data \cite{No_23}, as well as BiomedCLIP \cite{No_24}. Additionally, research on integrating ordinal relationships between classes in CLIP models has gained traction \cite{No_7,No_22}, offering promising applications for tasks requiring quantitative assessment in medical imaging.

\begin{figure}[t]
  \centering
  \includegraphics[width=1.0\linewidth]{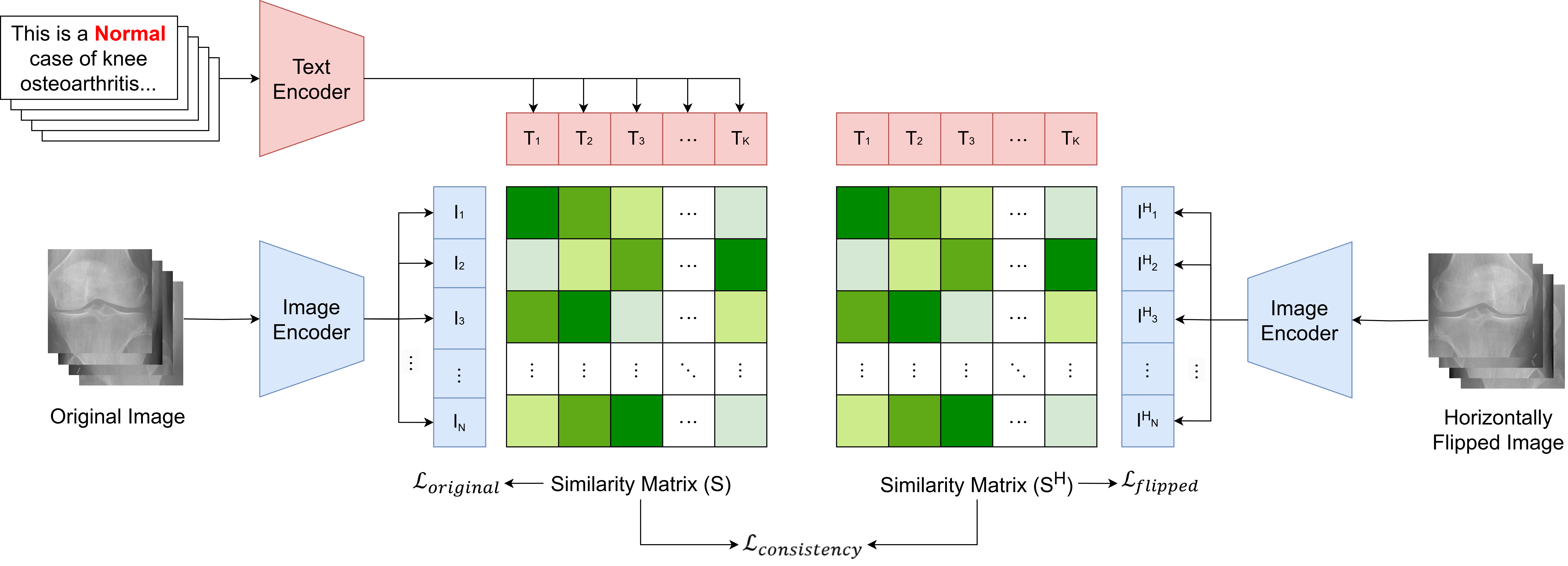}
  \caption{
  Symmetry Loss and Consistency Loss framework for knee osteoarthritis (KOA) severity grading. The original image $I$ and its horizontally flipped version $I^{H}$ are mapped to similarity matrices $\mathcal{S}$ and $\mathcal{S}^{H}$ through the image and text encoders. Symmetry Loss is computed using the cross-entropy loss between the similarity matrices and one-hot encoded labels. Jensen-Shannon Divergence (JSD) is used in $\mathcal{L}_{consistency}$ to measure the difference between the distributions of $\mathcal{S}$ and $\mathcal{S}^{H}$. The final loss function integrates Symmetry Loss and Consistency Loss to enhance the robustness of KOA severity classification.}
  \label{Figure_1}
\end{figure}

In this study, we propose a CLIP-based framework (CLIP-KOA) for the automatic and consistent classification of Knee Osteoarthritis (KOA). Given the high inter-observer variability and subjective nature of the existing Kellgren-Lawrence (KL) grading system, we introduce a novel learning approach that integrates both image and text modalities to address these limitations. 
The main contributions of this work are threefold. 
First, unlike existing KOA grading classification models that rely solely on image data, CLIP-KOA integrates both image and textual information to enable more sophisticated KOA severity prediction. 
This allows the model to go beyond simple KL grade (0–4) classification by learning both the meaning and features of each grade, leading to a more reliable KOA assessment. 
Second, we leverage the inherent symmetry of KOA images to design Symmetry Loss and Consistency Loss, which ensure prediction consistency between the original and flipped images. 
These loss functions encourage the model to produce consistent KOA ratings regardless of image orientation, thereby improving generalization performance. 
The schematic illustration of CLIP-KOA is shown in \cref{Figure_1}. 
Third, we empirically demonstrate that CLIP-KOA can predict KOA severity more precisely and consistently than existing CNN-based KOA grading models. 
While existing KL grading systems use only a fixed scale of 0–4, CLIP-KOA is designed to leverage textual information to capture more nuanced and subjective descriptions of KOA severity. 
This work suggests potential future research for overcoming the limitations of current KOA grading prediction methods with more refined multimodal image analysis approaches. 

\section{Methods}
\subsection{Problem formulation}
A knee osteoarthritis (KOA) image $I_{i}$ is provided as an input, with its corresponding severity grade $G_{i}$.
Thus, the dataset can be represented as a collection of paired samples: $\{I_{i}, G_{i}\} \in D$ where $D$ is the knee osteoarthritis dataset. 
The image encoder maps $I_{i}$ to an image embedding $x_{i}$, defined as : $\mathcal{X} = \{x_{i}\}^{N}_{i=1}$ where $N$ is the total number of images.
Given that there are $K$ distinct grades, we define the set of text embeddings as: $\mathcal{T} = \{t_{j}\}^{K}_{j=1}$.
The similarity matrix $\mathcal{S}$ is computed from the set $\mathcal{X}$ and $\mathcal{T}$ using cosine similarity: $\mathcal{S} = [s_{i,j}]_{N \times K}$,  $s_{i,j} = x_{i} \cdot t_{j}^T$.
The objective is to model a mapping function $f_{\theta} : I_{i} \xrightarrow{} G_{i}$, which maps an input knee osteoarthritis image $I_{i}$ to its corresponding grade $G_{i}$. This function is implemented as a deep neural network with learnable parameters $\theta$.

\subsection{Loss function}
The key idea of Symmetry Loss and Consistency Loss is that the model should maintain consistent predictions even when the left and right sides of the knee X-ray images are mixed. In other words, if the same knee osteoarthritis image is flipped left and right, the model should predict the same KL grade, and the prediction probability distribution should be similar.

\subsubsection{Symmetry Loss}
The main loss consists of both the image-text loss and horizontal flip image-text loss. 
Given an image $I_{i}$, its corresponding embedding vector $x_{i}$ and the text embedding vector $t_{j}$, the similarity score $s_{i,j}$ is computed as their inner product.
The similarity matrix $\mathcal{S} \in \mathbb{R}^{N \times K}$ represents the pairwise inner products between image and text embeddings.
Let $I_{i}^{H}$ be the horizontal flipped image of image $I_{i}$. Then, $\mathcal{S}^{H}$ represents the similarity matrix computed for the flipped image-text pairs. The label corresponding to $\mathcal{S}$ is denoted as $Y$ and is formulated as: 

\begin{align}
    Y = \mathbb{I}(S) = 
    \begin{cases}
        1 & \text{ground-truth grade} \\
        0 & \text{otherwise}
    \end{cases} 
    \; .
    \label{eq_1}
\end{align}

In the general CLIP framework, we use image-text pairs $(I_{i},T_{i})$ to learn the similarity between images and their corresponding text labels.
That is, it is optimized by increasing the similarity between the image and its correctly matched text pair $(I_{i},T_{i})$ and decreasing the similarity of the incorrect pair $(I_{i},T_{j})$.
However, since the goal of this study is to find labels that correctly match the text representing the corresponding rating of an image, the value of K is set to the number of grades defined in a given problem.
Therefore, the labels in Y are represented as one-hot encoding vectors corresponding to their respective grades.
Based on these rules, image-text loss and horizontal flipped image-text loss can be expressed as:

\begin{align}
    \mathcal{L}_{\text{symmetry}} = \mathcal{L}_{\text{original}} + \mathcal{L}_{\text{flipped}}
    = \text{CELoss}(S, Y) + \text{CELoss}(S^{H}, Y) \; .
\end{align}

\subsubsection{Consistency Loss}
While preserving the characteristics of a given knee osteoarthritis image $I$, the similarity score matrix $\mathcal{S}^H$ of the horizontally flipped image $I^H$ and the similarity score matrix $\mathcal{S}$ of the original image should share the same predictive probability distribution.
Therefore, Jensen-Shannon Divergence (JSD) \cite{No_26} is used to quantitatively evaluate the similarity between the two probability distributions \cref{eq_2}.

\begin{align}
    \mathcal{L}_{\text{consistency}} = JSD(P || Q) = \frac{1}{2} D_{KL}(P || M) + \frac{1}{2} D_{KL}(Q || M)  \; .
    \label{eq_2}
\end{align}

We define the predictive probability distribution of the original image as $P = \text{softmax}(\mathcal{S})$ and the predictive probability distribution of the horizontally flipped image as $Q = \text{softmax}(\mathcal{S}^{H})$, and set their mean as $M = \frac{1}{2} (P + Q)$.
This serves as a quantitative measure of the difference between two probability distributions $P$ and $Q$ by calculating the KL-Divergence based on the mean $M$ of the two distributions. Finally, the total loss function is defined as \cref{eq_3}.
Consistency Loss acts as a regularizer, encouraging the model to maintain consistent predictions across flipped images, ensuring that the same KOA image yields the same KL rating when flipped left and right.

\begin{align}
    \mathcal{L}_{\text{total}} = \frac{1}{2}(\mathcal{L}_{\text{original}} + \mathcal{L}_{\text{flipped}}) + \lambda \mathcal{L}_{\text{consistency}} \;,
    \label{eq_3}
\end{align}
where $\lambda$ is a hyperparameter, set to 10 by default, which defines relative weight of the consistency loss term $\mathcal{L}_{\text{consistency}}$ to the remaining portions of the loss.

\section{Experiments}

\subsubsection{Dataset}
We used the knee osteoarthritis (KOA) severity grading dataset to evaluate our method \cite{No_6}. The KOA dataset contains knee X-ray data for both knee joint detection and knee KL grading. The dataset consists of knee radiographs from 4796 participants.
All images were resized to 224 × 224 pixels to ensure consistency in resolution. The processed dataset included a total of 8260 images, which were split into training, validation, and test sets in a 7:1:2 ratio. \cref{table:Dataset} presents the distribution of the dataset across different KL grades.
For the text labels, we utilized the grading descriptions provided on Kaggle, which define the severity levels of knee osteoarthritis from Grade 0 to Grade 4. These descriptions serve as the standard criteria for KL grading, outlining the progression of osteoarthritis from healthy knees (Grade 0) to severe cases (Grade 4). Based on these descriptions, we generated text labels corresponding to each KL grade to enhance the interpretability and consistency of the dataset.

\begin{table}[t]
    \centering
    \caption{The distribution of knee osteoarthritis dataset}
    \setlength{\tabcolsep}{8pt}
    \begin{tabular}{ccccccc}
    \hline
               & Normal     & Doubtful     & Minimal     & Moderate     & Severe     & Total \\ \hline
    Train      & 2286       & 1046         & 1516        & 757          & 173        & 5578  \\
    Validation & 328        & 153          & 212         & 106          & 27         & 826   \\
    Test       & 639        & 296          & 447         & 223          & 51         & 1656  \\
    Total      & 3253       & 1495         & 2175        & 1086         & 251        & 8260  \\ \hline
    \end{tabular}
    \label{table:Dataset}
\end{table}

\subsubsection{Implementation Details}
Our model is built upon the ViT-B/16 Transformer architecture, utilizing a pre-trained CLIP backbone for both image and text encoding. The image encoder is based on a Vision Transformer (ViT), while the text encoder employs a masked self-attention Transformer. The implementation was carried out in PyTorch and trained on an NVIDIA RTX A6000 GPU.
During training, we employed the AdamW optimizer with a base learning rate of $1 \times 10^{-5}$ and a weight decay of $1 \times 10^{-6}$. The learning rate was adjusted using a OneCycleLR scheduler. We set the batch size to 64 for efficient training. Additionally, the knee joint images were normalized using the mean and standard deviation of the training dataset to ensure consistent input distributions.
To assess model performance, we evaluated classification results using accuracy, recall, precision, and F1-score, ensuring a comprehensive analysis of its effectiveness in knee osteoarthritis severity grading.

\section{Results}

\begin{table}[t]
    \centering
    \caption{ Comparison of accuracy across previous studies and the proposed CLIP-KOA model on the same dataset. }
    \setlength{\tabcolsep}{8pt}
    \begin{tabular}{clc}
    \hline
                                    & Method                & Accuracy         \\ \hline
    Chen et al. (2019)  \cite{No_14} & VGG 19 - Ordinal      & 69.6\%           \\
    Feng et al. (2021)  \cite{No_1}  & Attention module      & 70.23\%          \\
    Yong et al. (2022)  \cite{No_2}  & DenseNet 161 - ORM    & 70.23\%          \\
    Jain et al. (2024)  \cite{No_5}  & OsteoHRNET            & 71.74\%          \\
    \textbf{Ours}                   & \textbf{CLIP-KOA}     & \textbf{71.86\%} \\ \hline
    \end{tabular}
    \label{table:results_1}
\end{table}

In this study, we evaluated the performance of the proposed CLIP-KOA model in classifying the severity of knee osteoarthritis (KOA) and compared it with previous studies using the same dataset. As shown in \cref{table:results_1}, the comparison included models such as VGG 19 - Ordinal \cite{No_28}, Attention Module \cite{No_1}, DenseNet 161 - ORM \cite{No_2}, and OsteoHRNET \cite{No_5}. In the experimental results, the CLIP-KOA model outperformed the existing models, achieving an accuracy of 71.86\%, which is 0.12 percentage points higher than the previous best performer, OsteoHRNET (71.74\%).

Traditional CNN-based KOA classification models use only images as input and rely on extracting features from limited data. This approach makes it difficult to overcome data limitations, and performance tends to plateau around 70\%. In contrast, the CLIP-KOA model applies multimodal learning, incorporating textual data that includes detailed descriptions of grades rather than simply learning the grades themselves. This enables a more sophisticated differentiation of KOA severity.

In particular, the CLIP-KOA model is designed to go beyond the simple classification of KL grades (0–4) by learning the descriptions associated with each grade, allowing the model to develop a richer understanding of KOA severity. The existing KL Grading system consists of a fixed five-level scale (0–4), which has the limitation of not fully capturing the nuances of patients' actual symptoms and interpretations. However, CLIP-KOA utilizes textual keys to learn not only the numerical value of each grade but also its meaning and associated features, enabling a more detailed and precise KOA assessment. This approach is expected to contribute to the refinement of KOA grading and the development of a more sophisticated diagnostic system in the future.

\subsubsection{Ablation Study}
To verify the performance improvement of the CLIP-KOA model, we conducted an ablation study using the existing CLIP model. As shown in \cref{table:results_2}, the CLIP-KOA model achieved a 2.36\% increase in accuracy compared to the original CLIP model and demonstrated overall performance improvements in Precision, Recall, and F1-score.

\begin{figure}[t]
    \centering
    \begin{subfigure}[b]{0.48\textwidth}
        \centering
        \includegraphics[width=\textwidth]{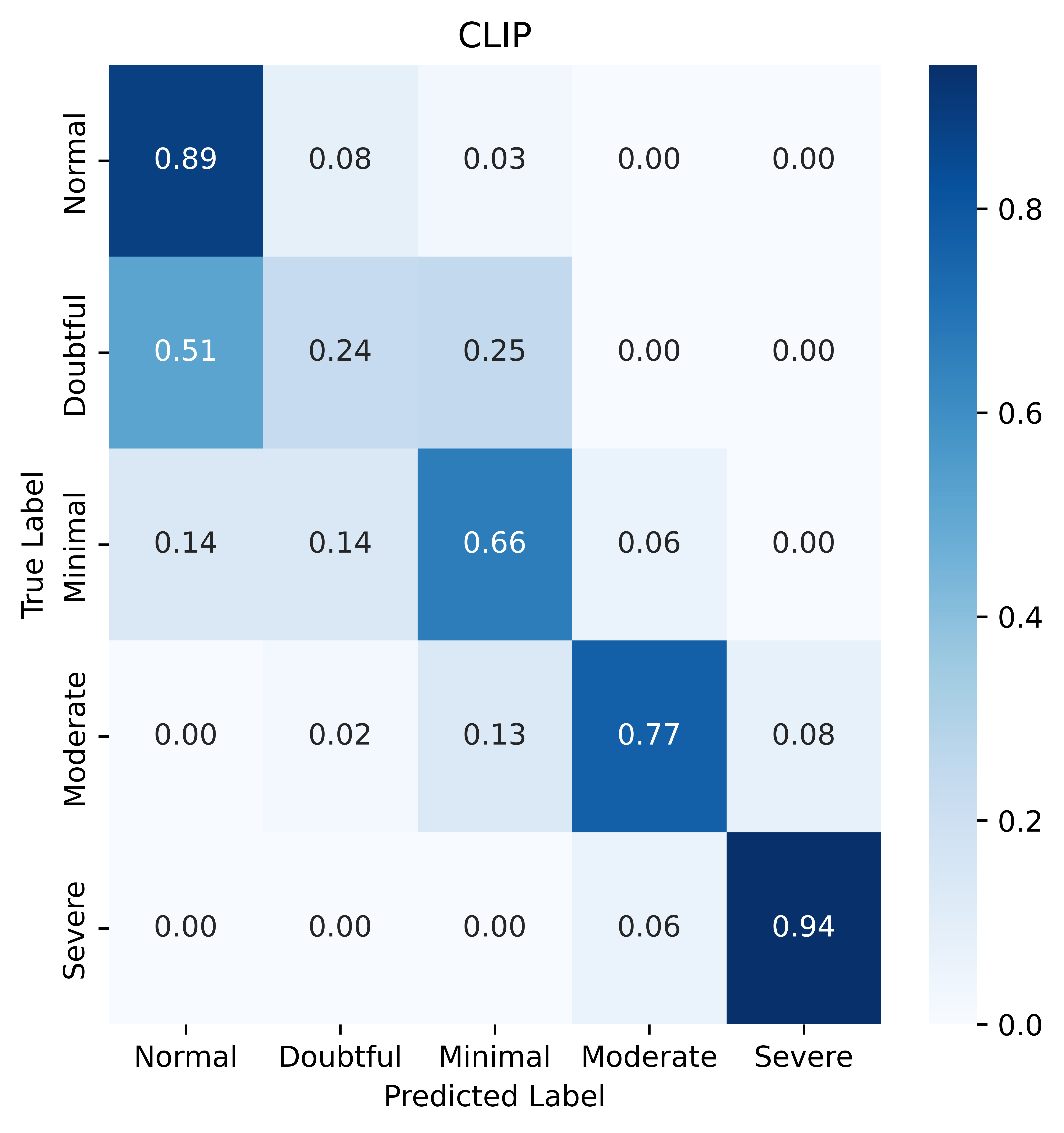}
    \end{subfigure}
    \hfill
    \begin{subfigure}[b]{0.48\textwidth}
        \centering
        \includegraphics[width=\textwidth]{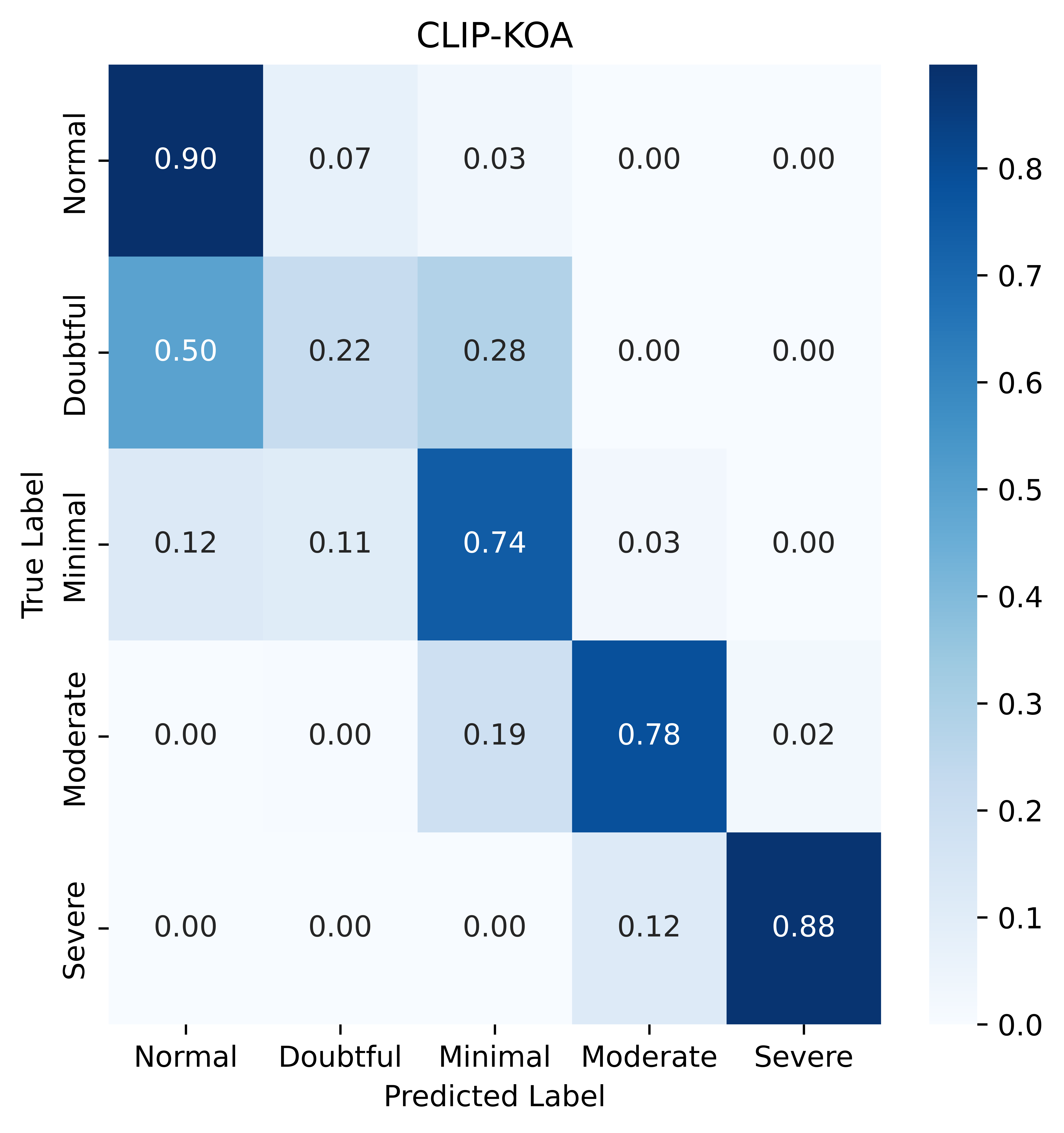}
    \end{subfigure}
    \caption{The figure presents a comparison of confusion matrices for KL grade prediction between the baseline CLIP model (left) and the proposed CLIP-KOA model (right). X-axis (Predicted Label): The five KL grades predicted by the model: Normal, Doubtful, Minimal, Moderate, and Severe. Y-axis (True Label): The actual KL grades assigned in the dataset.}
    \label{fig:results}
\end{figure}

Notably, the discrimination performance for intermediate KOA classes (KL 2–3) was significantly enhanced, with the prediction accuracy for the Minimal (KL 2) class increasing by 8\%, from 0.66 to 0.74, as shown in \cref{fig:results}. This suggests that the CLIP-KOA model effectively refines KOA grade prediction, particularly in distinguishing mild KOA from severe KOA, which is a crucial challenge in KOA classification.

These improvements can be attributed to the introduction of Symmetry Loss and Consistency Loss, which play a crucial role in refining the model’s predictive stability and robustness. Symmetry Loss ensures that the model maintains consistency when an image is horizontally flipped, enforcing robustness by aligning the predictions of original and flipped images. Meanwhile, Consistency Loss minimizes the divergence between the probability distributions of the original and flipped images, reinforcing stable predictions across different transformations.

By incorporating these loss functions, the CLIP-KOA model not only enhances classification accuracy but also ensures a more structured and reliable representation of KOA severity. This enables the model to generate more consistent predictions across varying image conditions, ultimately leading to improved differentiation of KOA severity grades and a more robust KOA classification system.

\begin{table}[t]
    \centering
    \caption{ Performance Comparison Based on Loss Application }
    \setlength{\tabcolsep}{8pt}
    \begin{tabular}{ccccc}
    \hline
                    & Accuracy (\%) & Precison & Recall & F1-Score \\ \hline
    CLIP            & 69.50         & 0.677    & 0.698  & 0.680    \\
    CLIP-KOA        & 71.86         & 0.729    & 0.705  & 0.709    \\ \hline
    \end{tabular}
    \label{table:results_2}
\end{table}

\section{Conclusion}

We propose a novel framework for predicting Kellgren and Lawrence (KL) grades from knee osteoarthritis (KOA) image data. Unlike previous studies that have primarily treated KL grade prediction as a regression or classification problem using only image-based models, our approach incorporates linguistic information to enhance prediction accuracy and consistency. Specifically, we leverage the structural symmetry of KOA images to ensure that original and flipped images yield consistent predictions, and introduce a multi-modal learning strategy that integrates textual descriptions related to KL grading.

The results of this study suggest that the CLIP-KOA model achieves competitive performance compared to existing CNN-based models in KOA severity classification. Unlike traditional models that rely solely on feature learning from image data, CLIP-KOA enhances KOA prediction accuracy and consistency by integrating a language model and introducing a new loss function. This multimodal approach enables a more structured and robust representation of KOA severity, particularly improving the discrimination of intermediate KL grades (KL 2–3).

Despite these advancements, challenges remain in the accurate prediction of early-stage KOA, particularly for the Doubtful (KL 1) rating, which was prone to misclassification. To further refine KOA severity classification, future research should explore more diverse textual inputs that provide richer descriptions of KOA symptoms and severity levels. Additionally, optimizing the loss function to incorporate adaptive weighting for specific KL grades could improve prediction accuracy, particularly in the early stages of KOA. Expanding the multimodal learning framework to integrate sequential patient data and clinical text reports may further enhance diagnostic reliability and model generalization.

\bibliographystyle{splncs04}
\bibliography{ref}

\end{document}